\documentclass[onecolumn]{PDS}
\usepackage{graphicx}
\usepackage{multicol,multirow}
\usepackage[fleqn]{amsmath}
\usepackage{amssymb,amsfonts}
\usepackage{mathrsfs}
\usepackage{amsthm}
\usepackage[figuresright]{rotating}
\usepackage{appendix}
\usepackage[authoryear]{natbib}
\usepackage{ifpdf}
\usepackage[T1]{fontenc}
\usepackage{textcomp}
\usepackage{xcolor}
\usepackage[colorlinks,allcolors=blue,bookmarks=false]{hyperref}
\usepackage{subfigure}
\usepackage[colorinlistoftodos]{todonotes}
\usepackage{wrapfig}
\usepackage{algorithm}
\usepackage{algpseudocode}

\jdoi{https://doi.org/10.1017/pds.2025.xxx                       }

\usepackage{boites}

\begin{document}

\articletype{AI and Data-Driven Design}%

\title{Towards Precision in Bolted Joint Design: 
\\A Preliminary Machine Learning-Based Parameter Prediction}
\author[1]{Ines Boujnah$^{*\dagger}$}
\author*[1]{Nehal Afifi$^{*\dagger}$}
\author[1]{Andreas Wettstein}
\author[1]{Sven Matthiesen}
\address[1]{IPEK – Institute of Product Engineering at the Karlsruhe Institute of Technology (KIT), Karlsruhe, Germany}
\renewcommand{\thefootnote}{\fnsymbol{footnote}}
\footnotetext[1]{These authors contributed equally to this work.}
\footnotetext[2]{ Joint first authorship.}

%\address[\textbf{$^{* \dagger}$}]{Joint first authorship.....}
%\address[2]{Second author address,}
%\address[3]{Third author address}
\corresemail{nehal.afifi@kit.edu}
\abstract{Bolted joints are critical in engineering for maintaining structural integrity and reliability. Accurate prediction of parameters influencing their function and behavior is essential for optimal performance. Traditional methods often fail to capture the non-linear behavior of bolted joints or require significant computational resources, limiting accuracy and efficiency. This study addresses these limitations by combining empirical data with a feed-forward neural network to predict load capacity and friction coefficients. Leveraging experimental data and systematic preprocessing, the model effectively captures nonlinear relationships, including rescaling output variables to address scale discrepancies, achieving 95.24\% predictive accuracy. While limited dataset size and diversity restrict generalizability, the findings demonstrate the potential of neural networks as a reliable, efficient alternative for bolted joint design. Future work will focus on expanding datasets and exploring hybrid modeling techniques to enhance applicability.}

%\abstract{Bolted joints are critical in technical applications and different everyday life applications, and accurate prediction of the functional parameters is essential. Traditional methods like calculations with VDI 2230, Finite Element Analysis (FEA), and Design of Experiments (DoE) struggle with capturing the nonlinear behavior of bolted joints, leading to a lack of reliable methods for analyzing load capacity and friction coefficients accurately and efficiently while accounting for nonlinearities. A supervised feedforward neural network was implemented as a data-driven approach to address these limitations, allowing for accurate predictions of the bolted joint’s nonlinear relationships without high computational demands. This model, trained on key parameters like bolt size, strength grade, preload force, tightening torque, head torque a thread torque, yielded an accuracy of 95.24\%.}


\maketitle

\section{Introduction} \label{sec:Introduction}

%As machinery's most common and versatile element \citep{spura_roloffmatek_2023}, bolted joints are universally applied across technical devices and various aspects of everyday life \citep{noauthor_schraubenverbindungen_2007}. Due to their versatility as detachable connections between two or more parts \citep{noauthor_schraubenverbindungen_2007, noauthor_vdi_2015}, they are used in applications from industry plants to bikes. These joints transmit loads between the connected parts \citep{noauthor_vdi_2015} and rely on torque application to either the head or the nut to create the preload force and with it the connection \citep{oberg_machinerys_2004}. 
%\citep{noauthor_schraubenverbindungen_2007, noauthor_vdi_2015}
Bolted joints, the most versatile machinery element \citep{spura_roloffmatek_2023}, are essential in technical devices and everyday applications, functioning as detachable connections between components \citep{noauthor_schraubenverbindungen_2007}. These joints facilitate load transmission between connected components \citep{noauthor_vdi_2015} by creating a preload force, achieved through the application of torque to either the head of the bolt or the nut \citep{oberg_machinerys_2004}. In the design process of bolted joints subjected to tensile loads, the preload force is a critical parameter influencing their functional behavior and reliability \citep{bickford_introduction_1995}. The friction coefficients in the head and thread regions of the bolt are equally significant, as the majority of the tightening torque is expended in overcoming friction in these areas \citep{noauthor_vdi_2024}. This friction not only contributes to the generation of preload force but also provides a self-locking effect, ensuring the joint remains secure under working conditions. As external loads are applied, the joint transitions through elastic and plastic deformation stages, potentially culminating in failure if critical load thresholds are exceeded \citep{steinhilper_konstruktionselemente_2012}. Poorly designed bolts may experience excessive stresses during tightening, increasing the risk of joint failure, equipment malfunction, or operational downtime \citep{noauthor_schraubenverbindungen_2007}. Thus, understanding the relationships between tightening variables like preload force and functional behavior parameters namely load-bearing capacity, and friction coefficients accurately and efficiently is essential for achieving a reliable design. 
\subsection{Related Work} \label{sec:Related Work}
%Bolted joint analysis traditionally relies on conventional methods like VDI 2230 calculations, FEA, and DoE to assess performance and predict behavior. Recently, data-driven approaches have emerged as a promising alternative, offering enhanced prediction accuracy by leveraging machine learning models. These new methods complement established techniques, providing improved insight into joint behavior under diverse conditions. 
%Bolted joint behavior has been analyzed through analytical, numerical, empirical, and data-driven methods
%%%%%%%%%%%%%%%%%%%%%%%%%%%%%%%%%%%%%%%%
%%%%%% intro & Analytical
Bolted joint behavior has traditionally been analyzed using a variety of modeling approaches. Analytical, numerical, empirical, and data-driven methods have all been extensively applied to investigate the behavior of bolted joints. \citep{noauthor_vdi_2015,young_fundamentals_2023}, each offering distinct strengths and facing inherent limitations. This section provides an overview of these approaches, highlighting their contributions and challenges, and identifies the research gap that motivates the proposed hybrid methodology. \textbf{Analytical methods} approximate bolted joint behavior through simplified mechanical representations, focusing on specific aspects of bolted joints. For instance, \citet{lee_methodology_2022} evaluates the tightening torque–clamping force relationship and friction coefficients, validating theoretical calculations through experimental measurements. \citet{zeng2020analytical} proposed a mathematical equation, derived from mechanical behavior, to predict the load-bearing capacity of beam-to-column joints in slim-floor composite frames. While analytical methods provide valuable insights into fundamental mechanics, they often lack the ability to capture the complex, non-linear interactions inherent in bolted joints. 
%Analytical, numerical, empirical, and data-driven methods have all been extensively applied to investigate the behavior of bolted joints. \citep{noauthor_vdi_2015,young_fundamentals_2023}, each offering distinct strengths and facing inherent limitations. This section provides a review of these approaches, highlighting their contributions and challenges, and identifies the research gap that motivates the proposed hybrid methodology. \textbf{Analytical models} simplify complex behaviors, focusing on specific aspects of bolted joints. For instance, \citet{lee_methodology_2022} developed a methodology to evaluate the tightening torque–clamping force relationship and friction coefficients, validating theoretical calculations through experimental measurements. While analytical methods provide valuable insights into fundamental mechanics, they often lack the ability to capture the complex, nonlinear interactions inherent in bolted joints. 

%%%%% Numerical Models %%%%%%   

To address these limitations, \textbf{numerical methods} such as finite element analysis (FEA), enable detailed representation of nonlinear material behavior, constraints, and complex geometries \citep{noauthor_vdi_2014}. They have been widely applied to study the mechanical behavior of bolted joints. \citet{fukuoka_mechanical_1998} focused on tension and torque characteristics during and after tightening through experimental and numerical methods. \citet{yu_finite_2015} explored how factors like friction, pitch, elastic modulus, and strain-hardening impact the tightening torque and initial load in bolted connections, using FEA to model these influences. Similarly, \citet{liang_investigation_2024} examined load capacity and fracture behavior in flange bolts under various load cases, providing insights into joint performance under different conditions. Although numerical approaches like FEA excel at modeling detailed mechanical behaviors, they often focus on idealized conditions, demand significant computational resources and rely heavily on the quality of the model and the expertise of the user to achieve accurate predictions \citep{noauthor_vdi_2014}.
%To address these limitations, \textbf{numerical approaches} such as finite element analysis (FEA) have been widely applied to study the mechanical behavior of bolted joints. 
%\citet{fukuoka_mechanical_1998} focused on tension and torque characteristics during and after tightening through experimental and numerical methods. \citet{yu_finite_2015} explored how factors like friction, pitch, elastic modulus, and strain-hardening impact the tightening torque and initial load in bolted connections, using FEA to model these influences. Similarly, \citet{liang_investigation_2024} examined load capacity and fracture behavior in flange bolts under various load cases, providing insights into joint performance under different conditions. Although numerical approaches like FEA excel at modeling detailed mechanical behaviors, they often focus on idealized conditions and require extensive computational resources. 
%%%%%%%%%%%%%%%%%%%%%%%%%%%%%%%%%%%%%%%
%%%%%%%%%%%% Empirical %%%%%%%%

To bridge the gap between simulation and real-world performance, \textbf{empirical methods} such as Design of Experiments (DoE) have been employed to systematically investigate factors influencing bolted joint behavior. \citet{nassar_experimental_2005, nassar_effect_2006} studied the effects of friction coefficients, tightening speed, and coating on the torque-tension relationship and wear patterns, while \citet{holch_evaluation_2024} examined load-bearing behavior in bolts and lockbolts under combined loading. While DoE offers a systematic approach to studying the effects of multiple factors on bolted joint behavior, and capable of accounting for nonlinearities to some degree, its implementation is not without challenges. The method requires carefully controlled and stable experimental conditions to ensure repeatability and reliability of results \citep{young_fundamentals_2023}. As the number of variables increases, the experimental design becomes increasingly complex, often demanding significant time and resources to execute \citep{selvamuthu_introduction_2024}. Additionally, external factors can introduce variability, reducing the robustness of the findings. The dependency of results on specific factors further restricts their applicability across different configurations \citep{wettstein_investigation_2020}. Although their widespread use, they face inherent challenges in accurately and efficiently predicting the functional behavior variables of bolted joints. Nonetheless, DoE remains a valuable method when used with other approaches to validate findings and enhance model accuracy \citep{afifi_2024}.

%To bridge the gap between simulation and real-world performance, \textbf{empirical methods} such as Design of Experiments (DoE) have been employed to systematically investigate factors influencing bolted joint behavior. \citet{nassar_experimental_2005, nassar_effect_2006} studied the effects of friction coefficients, tightening speed, and coating on the torque-tension relationship and wear patterns, while \citet{holch_evaluation_2024} examined load-bearing behavior in bolts and lockbolts under combined loading. While DoE offers a systematic approach to studying the effects of multiple factors on bolted joint behavior, its implementation is not without challenges. The method requires carefully controlled experimental conditions to ensure repeatability and reliability of results \citep{young_fundamentals_2023}. As the number of variables increases, the experimental design becomes increasingly complex, often demanding significant time and resources to execute \citep{selvamuthu_introduction_2024}. Additionally, external factors can introduce variability, reducing the robustness of the findings. The dependency of results on specific factors further restricts their applicability across different configurations \citep{wettstein_investigation_2020}. Despite these limitations, DoE remains a valuable method when used in conjunction with other approaches to validate findings and enhance model accuracy.
%%%%%%%%%%%%% Data-driven %%%%%

Furthermore, \textbf{data-driven predictive modeling} has emerged as a promising alternative to overcome these limitations. By leveraging relationships between input features and output variables, data-driven models can capture complex nonlinear behaviors and make  predictions about unseen scenarios, enabling more accurate and efficient predictions and optimizations of bolted joint performance \citep{montans_data-driven_2019}. For instance, \citet{fernandez-ceniceros_multilayer-perceptron_2012} used a neural network ensemble with FEA data to predict load capacity, while \citet{zhong_prediction_2021} combined FEA and neural networks for bearing capacity optimization in aluminum joints. \citet{fei_bolt_2016} modeled bolt force in flanges, analyzing the effects of bending and shear, and \citet{coelho_data-driven_2024} used Machine Learning (ML) to classify and quantify torque loss due to vibration. Other studies, such as those by \citet{yildirim_development_2019}, \citet{ren_prediction_2023}, and \citet{li_predicting_2024}, employed neural networks and ML algorithms to predict load capacity and other performance metrics in varied joint contexts, leveraging FEA and experimental data. \citet{chen_research_2020} and \citet{olejnik_friction_2023} further integrated optimization algorithms with neural networks for enhanced predictive precision, and \citet{atta_prediction_2019} focused on predicting failure stages in bolted joints with a neural network. This approach not only enhances the ability to predict critical parameters but also offers significant advantages in terms of time efficiency, accuracy, and ease of implementation. Yet they require high-quality and representative datasets for accurate training and validation. Recent research highlights the potential of combining empirical data with data-driven methods to address these challenges \citep{afifi_2024}. This study aims to explore this synergy, offering a framework that integrates experimental data with a feed-forward neural network model to enhance predictions of bolted joint performance. 

%Recent advancements in data-driven modeling have shown significant potential for addressing these challenges, enabling more accurate and efficient predictions and optimizations of bolted joint performance. For instance, citet{fernandez-ceniceros_multilayer-perceptron_2012} used a neural network ensemble with FEA data to predict load capacity, while \citet{zhong_prediction_2021} combined FEA and neural networks for bearing capacity optimization in aluminum joints.\citet{fei_bolt_2016} modeled bolt force in flanges, analyzing the effects of bending and shear, and \citet{coelho_data-driven_2024} used ML to classify and quantify torque loss due to vibration. Other studies, such as those by \citet{yildirim_development_2019, ren_prediction_2023, li_predicting_2024}, employed neural networks and ML algorithms to predict load capacity and other performance metrics in varied joint contexts, leveraging FEA and experimental data. \citet{chen_research_2020, olejnik_friction_2023} further integrated optimization algorithms with neural networks for enhanced predictive precision, and \citet{atta_prediction_2019} focused on predicting failure stages in bolted joints with a neural network.
\subsection{Problem Formulation and Task Description}
The reviewed studies illustrate the potential of data-driven modeling, including ML and neural networks, to advance the analysis of bolted joint. While these approaches offer accurate predictions of critical parameters, their integration with empirical data remains underexplored. A significant gap exists in methodologies that effectively combine the strengths of empirical experimentation and data-driven techniques to simultaneously predict load capacity and friction coefficients while capturing the nonlinear behavior of bolted joints. 
 %\subsection{Contributions} \label{sec:Contributions of this Paper}

This research aims to address this gap by leveraging the complementary advantages of empirical data and data-driven predictive modeling to accurately and efficiently predict critical function behavior parameters. 
It focuses on bolted joint behavior under tensile loading conditions under a torque-controlled tightening process, employing a supervised feed-forward neural network trained on key parameters, including bolt size, strength grade, tightening torque, head torque, thread torque, and preload force, this effectively models nonlinear relationships between inputs and outputs in a computationally efficient manner. 
This approach overcomes the limitations of traditional methods like analytical models, DoE, and FEA, which are hindered by idealized assumptions, scalability issues, and high computational demands. By integrating empirical data with predictive modeling, the study enhances accuracy, efficiency, and reliability, contributing to practical design applications.

%%%%%%%%%%% NN BACKGROUND %%%%%%%%%%%%
\section{Feedforward Neural Network Background} \label{sec:Feedforward Neural Network Background}

Data-driven predictive modeling aims to uncover relationships between input and output features, enabling accurate predictions on unseen data \citep{montans_data-driven_2019}. 
Within this field, ML focuses on developing algorithms capable of learning patterns from data to make informed predictions, with supervised ML specifically predicting target variables by analyzing input-output pairs  \citep{zhou_machine_2021,lecun_deep_2015}. 
Regression, as a key type of supervised learning problem, focuses on estimating numerical values based on input features \citep{Goodfellow-et-al-2016}. Linear regression handles simple relationships, whereas feed-forward neural networks handle complex patterns by processing information in a unidirectional flow, mapping inputs to outputs without cycles \citep{yadav_preliminaries_2015}. These networks feature a layered structure comprising an input layer, one or more hidden layers, and an output layer\citep{zhou_machine_2021}. Each layer consists of interconnected units called neurons, which represent input and output features \citep{douglas_j_santry_demystifying_2024,wythoff_backpropagation_1993}. The number of neurons in the hidden layers is determined by the network design and task complexity\citep{Goodfellow-et-al-2016}, enabling the processing of signals as outputs from one layer become inputs for the next \citep{douglas_j_santry_demystifying_2024, apicella_survey_2021}
Each layer's output is computed as the weighted sum of inputs plus a bias, with an activation function introducing non-linearity to enable complex mappings  \citep{kuhn_applied_2013}. Commonly used activation functions in neural networks include the sigmoid and Rectified Linear Unit (ReLU), while weight initialization methods often involve random or Xavier initialization \citep{douglas_j_santry_demystifying_2024}.

    In the context of optimization, the primary objective is to identify the optimal parameters of the network that minimize the loss function, which is a metric used to quantify the cost between the predicted values generated by the model and the actual target values \citep{montesinos_lopez_fundamentals_2022}. Gradient descent is one of the most commonly used algorithms for optimization. It works by iteratively updating the model's parameters in the direction opposite to the gradient until reaching a local or global minimum \citep{ruder2016overview}. A widely used variant is Stochastic Gradient Descent (SGD) \citep{ruder2016overview}, which computes the average gradient using one or more samples per epoch and updates all parameters based on this computation \citep{Ketkar2017}. The Huber loss represents one of the frequently utilized loss functions \citep{sadouk_robust_2020}.
    
            % The parameters of the network are optimized globally across the entire training set \citep{douglas_j_santry_demystifying_2024}, with updates occurring iteratively after each epoch \citep{Goodfellow-et-al-2016}. The success of this optimization is influenced by the initial values of weights and biases, which are typically initialized randomly \citep{douglas_j_santry_demystifying_2024}. As a result of this randomness, the process does not guarantee a global solution, meaning there is no assurance that the optimal result will be achieved \citep{kuhn_applied_2013}.
            
    % \begin{equation}
    %     x = x - \eta \cdot \nabla_x J(x; y; \widehat{y})
    %     \label{eq:sgd}
    % \end{equation}

    The principal objective of training a neural network is to fit its predicted outputs with the target outputs \citep{douglas_j_santry_demystifying_2024}. The testing phase evaluates the trained model's accuracy by generating predictions from the input data and calculating the error between the predicted and target outputs. \citep{douglas_j_santry_demystifying_2024, Goodfellow-et-al-2016}.

             % This process requires the input data to be provided to the network in discrete batches, with each batch comprising a specific number of data points \citep{Goodfellow-et-al-2016}. Subsequently, the loss between the predicted and target values is calculated, and the network's parameters are optimized to minimize this loss. This cycle of processing batches, computing loss, and updating parameters is repeated over numerous epochs until the loss reaches a satisfactorily low level \citep{douglas_j_santry_demystifying_2024}.
    
    Overfitting arises when a model performs exceptionally well on the training dataset but fails to generalize to unseen data, leading to poor performance on the test set. Conversely, underfitting occurs when a model performs poorly even on the training data set, suggesting that it has failed to capture the essential patterns in the data \citep{müller_introduction_2016}. Generalization is reached, when the error has been reduced to an adequately low level, thereby indicating that the model has acquired the capacity to make accurate predictions \citep{douglas_j_santry_demystifying_2024}.
           
\section{Methodology} \label{sec:Methodology}

\begin{figure}[ht]
    \centering{\includegraphics[width=0.95\textwidth]{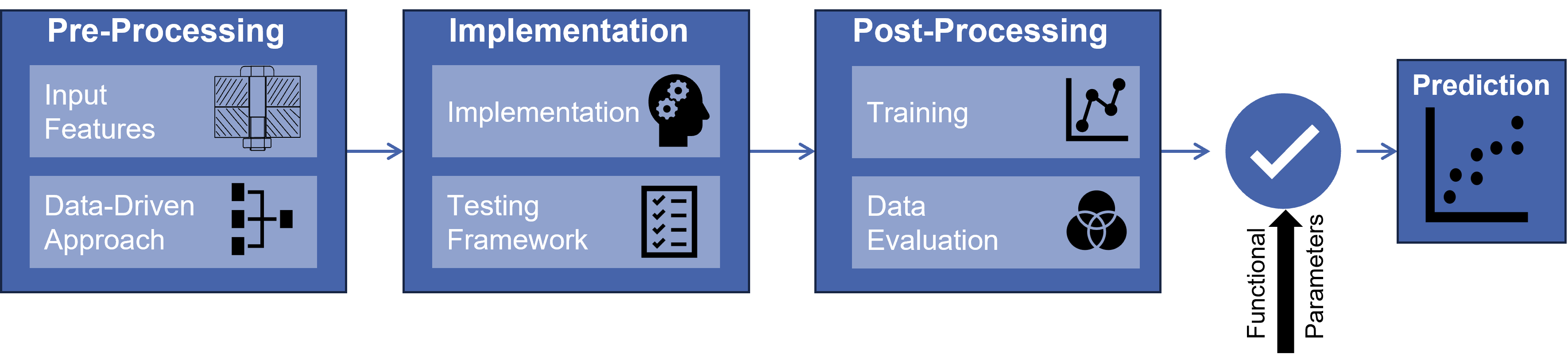}}
    \caption{Task Description\label{fig:task}}
\end{figure}

This section outlines the methodology for predicting key parameters in bolted joint design. Experimental data was processed and essential features were selected. A feed-forward neural network was implemented to model the complex relationships in bolted joint behavior. The fundamental methodology is illustrated in Figure \ref{fig:task}.

% \begin{wrapfigure}[25]{r}{0.3\textwidth}        
% \end{wrapfigure}

    \subsection{Data Description and Feature Selection} \label{sec:Data Description and Feature Selection}

%As, the friction at the bolt surface and the distribution of the load are influenced by bolt size, with larger bolts facilitating a more uniform load distribution and resulting in a larger friction generation. The strength grade of a bolt significantly impacts its load capacity, determining its ability to bear loads and how stress is distributed throughout the material. The tightening torque can be divided into two distinct components: the head torque and the thread torque. The magnitude of these torques is directly proportional to the friction coefficients. The remaining tightening torque generates the preload force. The magnitude of the torque applied influences the amount of load introduced to the bolt \citep{noauthor_vdi_2024}. So our selected input parameters are the bolt size, strength grade, preload force, tightening torque, head torque, and thread torque. and the output parameters are the parameters that affect the functional behavior of the bolted joint namely the load capacity and the head and the thread friction coefficients.
To establish a rationale for the chosen data, it is essential to consider the factors influencing bolted joint performance. The friction at the bolt surface and the load distribution are influenced by the bolt size, with larger bolts promoting a more uniform load distribution and greater friction generation \citep{noauthor_vdi_2015}. Additionally, the strength grade of a bolt plays a crucial role in determining its load capacity and how stress is distributed throughout the material \citep{noauthor_schraubenverbindungen_2007}. Furthermore, head torque and thread torque are both directly proportional to the friction coefficients. The remaining tightening torque contributes to the generation of preload force, with the applied torque magnitude directly affecting the load introduced to the bolt \citep{noauthor_vdi_2024}. Based on these considerations, the selected input parameters are bolt size, strength grade, preload force, tightening torque, head torque, and thread torque, while the output parameters are those influencing the functional behavior of the bolted joint, specifically load capacity and head and thread friction coefficients.

The experimental data were obtained from two sets of controls, each involving the continuous tightening of a bolted joint consisting of a bolt, two connected plates, a washer, and a nut. The initial dataset was gathered for an M10 8.8 ISO 4014 bolted connection, as described by \citet{wettstein_investigation_2020}. The second dataset involved an M6 8.8 ISO 4017 bolted joint, obtained with the same methodology. Time series data consisting of preload force, tightening torque, head torque, and thread torque were measured. For the M6 bolt, a preload force of 8 kN was used, with 20 samples collected. The M10 model utilized a preload force of 12.5 kN, with 9 samples available, while the M10 bolt with a preload force of 25 kN included 5 samples. From these values, the friction coefficients of the head and threads and the remaining load capacity were further empirically estimated. The set finally had 34 samples for training and testing the neural network. For the first 79 models, a reduced data size was used with 28 samples. 
%The selected input parameters were determined as the bolt size strength grade, preload force, tightening torque, head torque, and thread torque. Furthermore, the output parameters are the functional behavior parameters namely; the load capacity and the friction coefficients.

 % \citep{iso4017}  \citep{iso4014}

% \begin{table}[h]
%     \centering
%     \caption{Data\label{tab:data}}
%     \begin{tabular*}{\textwidth}{@{\extracolsep{\fill}}lllll@{}}
%         \toprule
%          & M6 & \multicolumn{2}{c}{M10} \\
%         \midrule
%         Preload Force [kN] & 8 & 12.5 & 25\\
%         Number of Samples & 20 & 9 & 5 \\
%         \bottomrule
%     \end{tabular*}
% \end{table}
           
    \subsection{Data Preprocessing} \label{sec:Data Preprocessing}
    
        The data provided by \citet{wettstein_investigation_2020} was presented in tabular format, representing time series measurements of the variables. The dataset was divided into two subsets: a training (80\%) and a test (20\%) set. Input and output features were subsequently extracted according to predefined feature specifications. When scaling was required, the input features were transformed through normalization or standardization, depending on the chosen method (Section \ref{sec:Model Architecture}). The scale derived from the training set was retained to ensure consistent application during subsequent stages of analysis.
        The preprocessing procedures applied to the test set mirrored those of the training set, with a critical distinction: any scaling of the test data was performed using the parameters obtained from the training set. This approach ensured that the model operated on data with a consistent scale, thereby preserving the validity of the training and evaluation processes.
   
    \subsection{Model Architecture} \label{sec:Model Architecture}

\begin{figure}[ht]
            \centering{\includegraphics[width=0.55\textwidth]{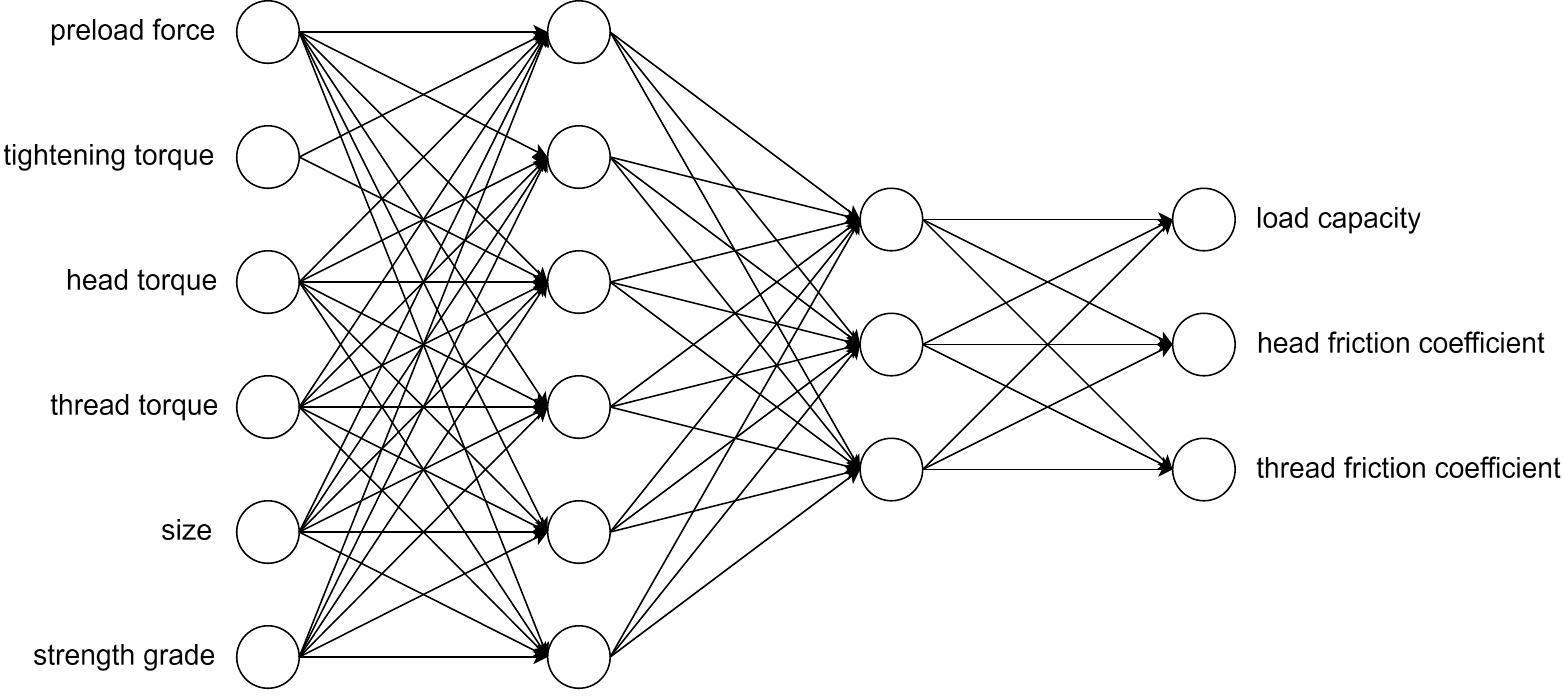}}
            \caption{Visualization of the Used Network Architecture\label{fig:network architecture}}
        \end{figure}

        The input layer contained six nodes, representing bolt size, strength grade, tightening torque, head torque, thread torque, and preload force. The neural network included two hidden layers, with neuron counts matching the input and output layers. The network's output layer was fixed with three nodes, corresponding to the output features: load capacity, head friction coefficient, and thread friction coefficient (Figure \ref{fig:network architecture}).

        % \begin{wrapfigure}{r}{0.6\textwidth}        
        % \end{wrapfigure}

        \begin{table}[ht]
            \centering
            \caption{Hyperparameters of the Model\label{tab:model hyperparameters}}
            \begin{tabular*}{\textwidth}{@{\extracolsep{\fill}}lllll@{}}
                \toprule
                & Optimization Algorithm & Learning Rate &
                Loss Function & Batch Size\\
                \midrule
                Hyperparameters & SGD & 0.01 & Huber & 4 \\
                \bottomrule
            \end{tabular*}
        \end{table}
        
        The model was trained using SGD with a learning rate of 0.01, aiming to optimize convergence speed and minimize loss. The Huber loss function was applied to balance sensitivity to outliers and improve stability. The training was conducted with a batch size of 4, enabling incremental updates to the model’s parameters (Table \ref{tab:model hyperparameters}).

        A total of 136 models were trained throughout the process. Four representative models were selected to illustrate the learning and optimization process. These models were evaluated with variations in key hyperparameters, including the activation function, weight initialization method, number of epochs, scaling method, units for preload force and load capacity, and the number of samples used (Table \ref{tab:hyperparameters}). The activation function determined the scale of outputs, while the weight initialization method affected training speed, convergence, and final accuracy. The number of epochs ensured sufficient training to generalize without overfitting. The scaling method aligned the data within an optimal range, reducing outliers and improving convergence. Choosing appropriate units for preload force and load capacity ensured output compatibility, and a larger dataset size enhanced pattern recognition.
        
        \begin{table}[ht]
            \centering
            \caption{Hyperparameters of Chosen Models\label{tab:hyperparameters}}
            \begin{tabular*}{\textwidth}{@{\extracolsep{\fill}}lllll@{}}
                \toprule
                 Hyperparameters & Model 1 & Model 2 &
                Model 3 & Model 4\\
                \midrule
                Activation Function & Sigmoid & ReLU & Sigmoid & Sigmoid \\
                Initialization Method & Random & Random & Xavier & Xavier \\
                Number of Epochs & 1000 & 1000 & 5500 & 4800 \\
                Scaling & Standardization & Standardization & Normalization & Normalization \\
                Unit [Preload/Load] & N/N & kN/kN & kN/kN & kN/MN \\
                Number of Samples & 28 & 28 & 28 & 34 \\
                \bottomrule
            \end{tabular*}
        \end{table}

        % \begin{figure}[!ht] 
        %     \centering{\includegraphics[width=0.5\textwidth, height=0.65\textheight]{pictures/pseudocode.pdf}}
        %     \caption{Flowchart of the Implemented Program\label{fig:pseudocode}}
        % \end{figure}
        
         \textbf{The initialization stage} involved configuring all hyperparameters prior to the training process, establishing the foundation for learning. These included the activation function, initialization method for weights and biases, network width and depth, proportion of data used for training and testing, number of epochs, batch size, loss function, optimization algorithm, and learning rate. The neural network structure was created by defining the total number of layers and the number of nodes per layer. The weights were initialized following the selected method, with random initialization serving as the default. If a particular technique was required, it was assigned following the activation function utilized. The Xavier initialization was applied to the sigmoid activation functions, whereas the He initialization was utilized for ReLU. The bias was initialized in one of two ways: randomly or to a constant value of zero for each neuron.
        
    % \subsection{Training Framework} \label{sec:Training Framework} 

\begin{figure}[ht] 
                \centering{\includegraphics[width=0.9\textwidth]{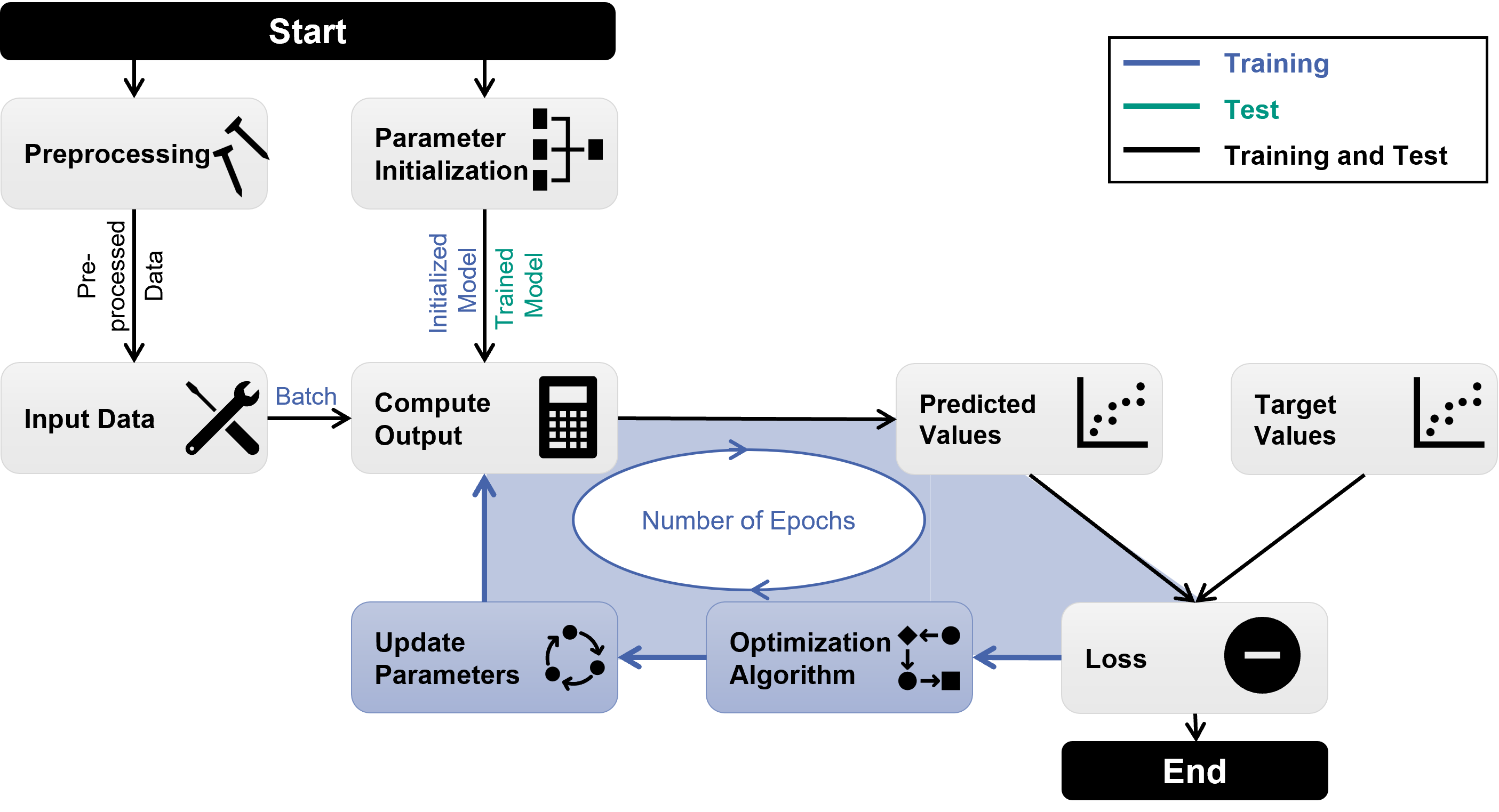}}
                \caption{Visualization of the Implemented Workflow\label{fig:loop}}
            \end{figure}

        \textbf{The training process} commenced following the initialization of the neural network. The training dataset was loaded and shuffled to ensure varied data processing during each iteration, improving efficiency. The loss function, optimization algorithm, and learning rate were then configured. Over a specified number of epochs, the training loop iterated with a defined batch size, through the data. The network generated outputs from the input data in each iteration, calculated the loss, and updated learnable parameters based on the chosen optimization algorithm. Mean accuracy and loss values were computed and plotted after each epoch to track progress and evaluate performance throughout training. This process is described in Figure \ref{fig:loop} by the blue and black path and by Algorithm \ref{alg:training}.
\vspace{-0.3cm}
        \begin{algorithm}
        \caption{Training Framework}\label{alg:training}
        \begin{algorithmic}
        \For{$epoch < number\_of\_epochs$}
        \For{$batch < batch\_size$}
            \State $optimizer.zero\_grad()$
            \State $predicted\_output \gets model(batch\_input)$
            \State $loss \gets loss\_function(predicted\_output, batch\_output)$
            \State Call $loss.backward()$
            \State Call $optimizer.step()$
        \EndFor
        \EndFor
        \end{algorithmic}
        \end{algorithm}
  \vspace{-0.3cm}      
    % \subsection{Testing Framework} \label{sec:Testing Framework}

         \textbf{The testing process} involved loading the unshuffled test dataset for evaluation, limited to a single epoch and batch due to the smaller size of the test set, ensuring computational efficiency. The neural network computed outputs based on the input data and model accuracy was assessed for each data point, with the mean accuracy calculated for the overall model accuracy. Predictions were deemed accurate if they deviated by no more than 5\% from the target values. To evaluate performance, error metrics such as Mean Absolute Error (MAE), Mean Square Error (MSE), and Root Mean Square Error (RMSE) were computed for each data point. A scatter plot was also generated, displaying predicted outputs against target values, where accurate predictions aligned along a linear function through the origin. This process is described in Figure \ref{fig:loop} by the green and black path and by Algorithm \ref{alg:testing}.
\vspace{-0.3cm}
        \begin{algorithm}
        \caption{Testing Framework}\label{alg:testing}
        \begin{algorithmic}
        \State With $torch.no\_grad$
        \For{$batch < batch\_size$}
            \State $model.eval()$
            \State $predicted\_output \gets model(batch\_input)$
        \EndFor
        \end{algorithmic}
        \end{algorithm}
 \vspace{-0.3cm}                
\vspace{-0.2cm}
    % \subsection{Prediction Framework} \label{sec:Prediction Framework}
    
    %     Following the conclusion of its training, the model was suitable for immediate deployment in a predictive context. To accomplish this, the trained model was loaded, and the structure with which it was trained was initialized. Subsequently, the parameters were updated with the values derived from the trained model, enabling the model to make accurate predictions based on new input data.
        
    % \subsection{Evaluation Metrics} \label{sec:Evaluation Metrics}

        \textbf{The evaluation and hyperparameter tuning process} involved analyzing the accuracy and loss curves from training, as well as the error metrics and accuracy curves from testing, to assess the model's alignment with the data. Hyperparameters were iteratively adjusted until satisfactory accuracy was achieved. Once optimized, the trained model parameters were saved for future use.
        
\section{Results} \label{sec:Results}

\begin{figure}[!ht]    
        \centering{\includegraphics[width=0.53\textwidth]{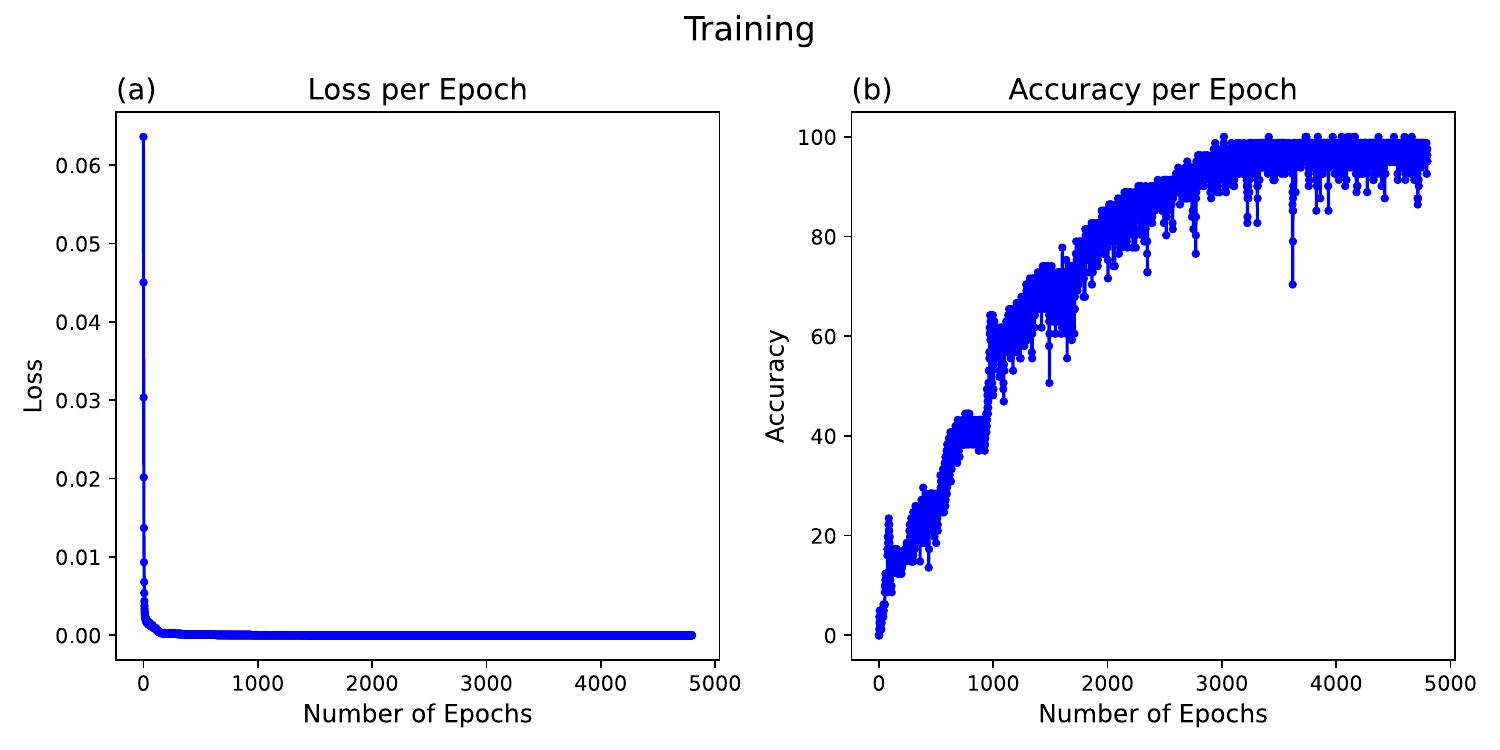}}
        \caption{Results of Model 4: Training (a) Loss per Epoch (b) Accuracy per Epoch\label{fig:training}}
    \end{figure}

\begin{figure}[!ht]
        \centering{\includegraphics[width=0.8\textwidth]{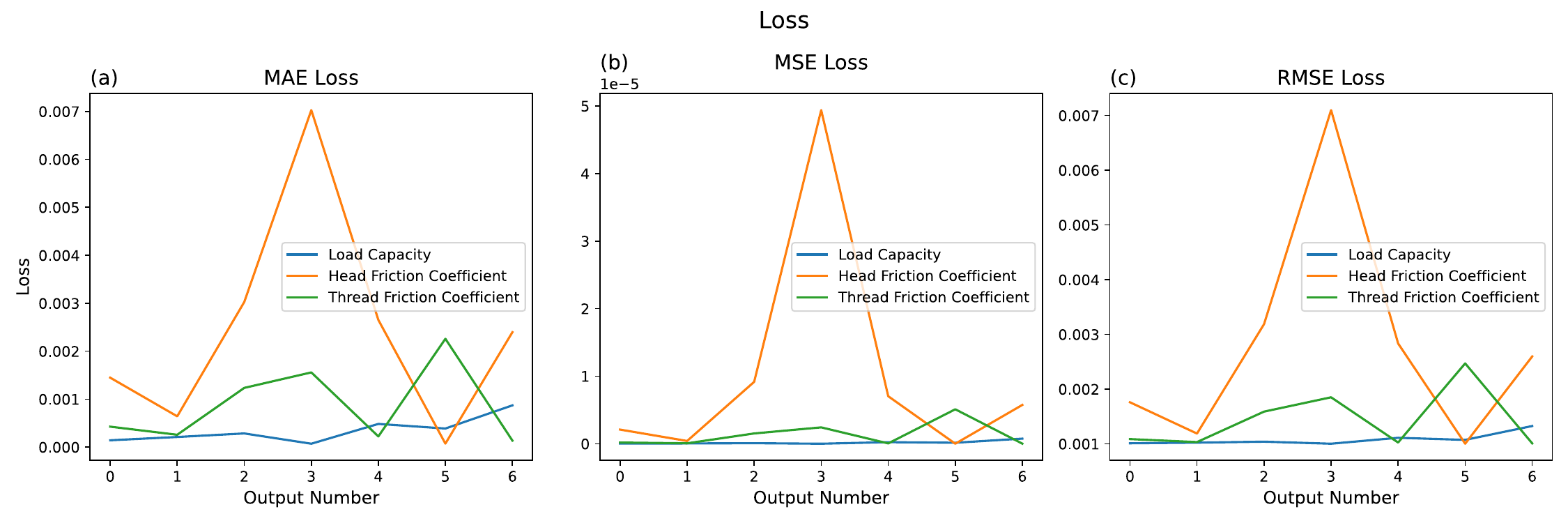}}
        \caption{Results of the Model 4: Loss (a) MAE Loss (b) MSE Loss (c) RMSE Loss\label{fig:loss}}
    \end{figure}

\begin{figure}[!ht]
        \centering{\includegraphics[width=0.8\textwidth]{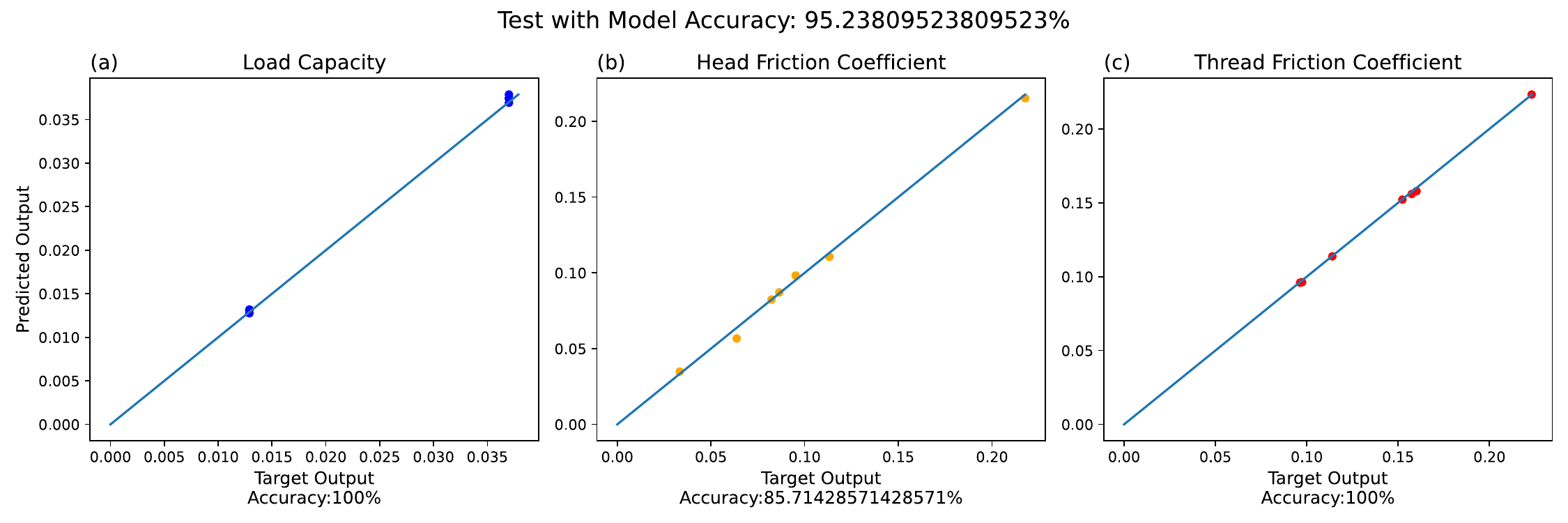}}
        \caption{Results of Model 4: Test (a) Scatter of the Load Capacity (b) Scatter of the Head Friction Coefficients (c) Scatter of the Thread Friction Coefficients\label{fig:test}}
    \end{figure}

    As the hyperparameters tuning process progressed, the accuracy of each model demonstrated an incremental improvement, accompanied by an increase in the elapsed time required for completing all steps, from preprocessing to training. The final accuracy attained for Model 4 was 95.24\%, with a generation time of 90 seconds (Table \ref{tab:results}).
    After the hyperparameters tuning, the training curves followed the typical pattern for the loss and accuracy, with the loss decreasing logarithmically almost to zero and the accuracy increasing exponentially to a maximum of around 100\% over the epochs (Figure \ref{fig:training}).  
     The loss between the predicted and actual values was nearly negligible for all metrics (MAE, MSE, RMSE), except one of the head friction coefficient values, which exhibited a considerable deviation from the other values (Figure \ref{fig:loss}).
     In Figure \ref{fig:test} the predicted values for all output parameters aligned closely with the reference line through the origin for the scatter plot of the predicted to the target values. The load capacity and thread friction coefficients achieved 100\% accuracy, while the head friction coefficients reached 85.71\%, with only one value showing a minimal deviation from the line. All other values are within the 5\% range of the blue line.
\vspace{-0.3cm}    
    \begin{table}[ht]
        \centering
        \caption{Results of Chosen Models\label{tab:results}}
        \begin{tabular*}{\textwidth}{@{\extracolsep{\fill}}lllll@{}}
            \toprule
             & Model 1 & Model 2 &
            Model 3 & Model 4\\
            \midrule
            Accuracy [$\%$] & 11.11 & 33.33 & 66.67 & 95.24 \\
            Elapsed Time [s] & 2.3738 & 10.4265 & 5.4727 & 90.1053 \\
            \bottomrule
        \end{tabular*}
    \end{table}
    \vspace{-0.6cm}

\section{Discussion} \label{sec:Discussion} 

This study achieves significant improvements in predicting load capacity and friction coefficients in bolted joints by addressing challenges in model architecture and preprocessing. A primary challenge was the scale discrepancy between the output variables, specifically load capacity and friction coefficients. Rescaling the load capacity to MN effectively aligned the scales, minimized noise in the loss function, and enhanced the stability of the training process. Furthermore, the adoption of the sigmoid activation function, appropriately matched to the range of the output variables, facilitated balanced prediction accuracy across both parameters. These results underscore the importance of systematic preprocessing and model optimization in developing robust predictive tools for complex engineering systems. 
%The approach developed in this study led to significant improvements in predicting the load capacity and friction coefficients of bolted joints. By systematically addressing key challenges in data preprocessing and model architecture, the proposed neural network model achieved high accuracy and enhanced reliability in predictions. A major challenge encountered was the scale discrepancy between the output variables—load capacity (measured in N or kN) and friction coefficients (ranging from 0 to 1). To mitigate this issue, we rescaled the load capacity to MN, thereby aligning the scales of the output variables and reducing the noise in the loss function during training. Furthermore, we compared several activation functions and selected sigmoid, as it was more compatible with the range of the output variables. This choice not only improved the accuracy of load capacity predictions, but also facilitated a better balance between the accuracy of load capacity and friction coefficient predictions.

The dataset used in this study, consisting of only 28 to 34 samples, presents limitations in terms of generalizability and the stability of results across different bolt configurations. While the model demonstrated high accuracy for certain configurations, the limited diversity of the dataset raises concerns about overfitting, and the performance should, therefore, be interpreted with caution. To improve the generalizability of the model, future work will focus on expanding the data set to include a wider variety of bolt configurations and operational conditions, ensuring that the model can make reliable predictions for bolts in active use. Furthermore, we plan to explore techniques to generate synthetic data and augment the dataset through rule-based formulations of bolt behavior like VDI \cite{noauthor_vdi_2015} or FEA \cite{noauthor_vdi_2014}. This data-driven approach could prove valuable, as the acquisition of empirical data is both costly and time-consuming.

Finally, we believe that the accessibility of the model's design enables users to achieve accurate predictions without requiring advanced mechanical expertise. This versatility, coupled with the model's potential for time-efficient prediction of output variables, underscores its practical application. Nevertheless, expanding the dataset remains a critical next step to improve prediction consistency, generalizability, and robustness across a broader range of bolted joint configurations.

\section{Conclusion} \label{sec:Conclusion}
This study presented a hybrid approach integrating empirical data with a supervised feed-forward neural network to predict critical performance metrics of bolted joints, specifically load capacity and head and thread friction coefficients. The proposed methodology effectively addresses key limitations of traditional approaches such as analytical simplifications, computational demands of numerical models, and the scalability challenges of empirical methods. By leveraging empirical data and a computationally efficient neural network architecture, the model captured the nonlinear relationships between input parameters, such as tightening torque and preload force, and output parameters namely remaining load-bearing capacity and friction coefficients with a predictive accuracy of 95.24\%.
The findings highlight the potential of data-driven modeling as a transformative tool for bolted joint design, particularly in its ability to combine experimental rigor with advanced predictive capabilities. However, limitations such as the small dataset size and the reliance on simplified loading conditions pose challenges to the model’s generalizability across diverse scenarios. These limitations suggest the need for future research to expand the dataset, incorporate varied load cases (e.g., shear and combined loading), and explore alternative neural network architectures or hybrid machine learning models to enhance robustness and accuracy. 
This research bridges empirical experimentation and predictive modeling, offering a novel approach to bolted joint analysis. It integrates advanced computational techniques into engineering design, enabling more reliable and efficient optimization of bolted joint performance.
\begin{Backmatter}
\section*{Acknowledgements}
This work was funded by the Deutsche Forschungsgemeinschaft (DFG, German Research Foundation) – "Functional modeling of bolted joints under uncertain product conditions for remanufacturing" - with the project number 525034540 
\bibliographystyle{apalike}
\bibliography{PDS}

\begin{thebibliography}{}

\bibitem[Afifi et~al., 2024]{afifi_2024}
Afifi, N., Kaiser, J.-P., Andreas, W., Gisela, L., and Matthiesen, S. (2024).
\newblock Data-driven functional modeling of corroded bolted joints: A methodological framework for remanufacturing.
\newblock In {\em Proceedings of the ASME 2024 International Mechanical Engineering Congress and Exposition (IMECE2024)}, Portland, Oregon.

\bibitem[Apicella et~al., 2021]{apicella_survey_2021}
Apicella, A., Donnarumma, F., Isgr{\`o}, F., and Prevete, R. (2021).
\newblock A survey on modern trainable activation functions.
\newblock {\em Neural Networks}, 138:14--32.

\bibitem[Atta et~al., 2019]{atta_prediction_2019}
Atta, M., Abd-Elhady, A., Abu-Sinna, A., and Sallam, H. (2019).
\newblock Prediction of failure stages for double lap joints using finite element analysis and artificial neural networks.
\newblock {\em Engineering Failure Analysis}, 97:242--257.

\bibitem[Bengio et~al., 2016]{Goodfellow-et-al-2016}
Bengio, Y., Goodfellow, I., and Courville, A. (2016).
\newblock {\em Deep learning}, volume~1.
\newblock MIT press Cambridge, MA, USA.

\bibitem[Bickford, 1995]{bickford_introduction_1995}
Bickford, J. (1995).
\newblock An introduction to the design and behavior of bolted joint marcel dekker.
\newblock {\em New York}, 894:41.

\bibitem[Chen et~al., 2020]{chen_research_2020}
Chen, Y., Zhang, J., Liu, Y., Zhao, S., Zhou, S., and Chen, J. (2020).
\newblock Research on the prediction method of ultimate bearing capacity of pbl based on iaga-bpnn algorithm.
\newblock {\em IEEE Access}, 8:179141--179155.

\bibitem[Chiang and Chiang, 2023]{young_fundamentals_2023}
Chiang, Y.~J. and Chiang, A.~L. (2023).
\newblock {\em Fundamentals of Design of Experiments for Automotive Engineering Volume I}, volume~1.
\newblock SAE International.

\bibitem[Coelho et~al., 2024]{coelho_data-driven_2024}
Coelho, J.~S., Machado, M.~R., Dutkiewicz, M., and O.~Teloli, R. (2024).
\newblock Data-driven machine learning for pattern recognition and detection of loosening torque in bolted joints.
\newblock {\em Journal of the Brazilian Society of Mechanical Sciences and Engineering}, 46(2):75.

\bibitem[Fei et~al., 2016]{fei_bolt_2016}
Fei, Y., Pengdong, G., and Yongquan, L. (2016).
\newblock Bolt force prediction using simplified finite element model and back propagation neural networks.
\newblock In {\em 2016 IEEE Information Technology, Networking, Electronic and Automation Control Conference}, pages 520--523. IEEE.

\bibitem[Fern{\'a}ndez-Ceniceros et~al., 2012]{fernandez-ceniceros_multilayer-perceptron_2012}
Fern{\'a}ndez-Ceniceros, J., Sanz-Garc{\'\i}a, A., Anto{\~n}anzas-Torres, F., and Mart{\'\i}nez-de Pis{\'o}n-Ascacibar, F.~J. (2012).
\newblock Multilayer-perceptron network ensemble modeling with genetic algorithms for the capacity of bolted lap joint.
\newblock In {\em Hybrid Artificial Intelligent Systems: 7th International Conference, HAIS 2012, Salamanca, Spain, March 28-30th, 2012. Proceedings, Part I 7}, pages 545--556. Springer.

\bibitem[Fukuoka and Takaki, 1998]{fukuoka_mechanical_1998}
Fukuoka, T. and Takaki, T. (1998).
\newblock Mechanical behaviors of bolted joint during tightening using torque control.
\newblock {\em JSME International Journal Series A Solid Mechanics and Material Engineering}, 41(2):185--191.

\bibitem[Holch et~al., 2023]{holch_evaluation_2024}
Holch, A., Glienke, R., D{\"o}rre, M., and Henkel, K. (2023).
\newblock Evaluation of the load-bearing behaviour of bolts and lockbolt systems under combined tension and shear loading.
\newblock In {\em International Conference on Advanced Joining Processes}, pages 3--13. Springer.

\bibitem[Ketkar and Santana, 2017]{Ketkar2017}
Ketkar, N. and Santana, E. (2017).
\newblock {\em Deep learning with Python}, volume~1.
\newblock Springer.

\bibitem[Kloos and Thomala, 2007]{noauthor_schraubenverbindungen_2007}
Kloos, K. and Thomala, W. (2007).
\newblock {\em Schraubenverbindungen}.
\newblock Springer.

\bibitem[Kuhn, 2013]{kuhn_applied_2013}
Kuhn, M. (2013).
\newblock {\em Applied predictive modeling}.
\newblock Springer.

\bibitem[LeCun et~al., 2015]{lecun_deep_2015}
LeCun, Y., Bengio, Y., and Hinton, G. (2015).
\newblock Deep learning.
\newblock {\em nature}, 521(7553):436--444.

\bibitem[Lee et~al., 2022]{lee_methodology_2022}
Lee, J., Kim, D., and Seok, C.-S. (2022).
\newblock Methodology for evaluating the tightening torque-clamping force relationship and friction coefficients in bolted joints.
\newblock {\em Journal of Mechanical Science and Technology}, 36(4).

\bibitem[Li et~al., 2024]{li_predicting_2024}
Li, X., Liu, C., Xue, Y., Xue, S., Liao, S., and Zhou, Y. (2024).
\newblock Predicting bearing capacity of angle steel bolted connections using machine learning based on experimental and numerical database.
\newblock In {\em Structures 67}. Elsevier.

\bibitem[Liang et~al., 2024]{liang_investigation_2024}
Liang, K., Wang, Z., Yin, Z., and Hao, P. (2024).
\newblock Investigation on bearing capacity and fracture behaviour of the bolt in a flange connection considering multiple load cases.
\newblock In {\em Structures}, volume~62, page 106252. Elsevier.

\bibitem[Mont{\'a}ns et~al., 2019]{montans_data-driven_2019}
Mont{\'a}ns, F.~J., Chinesta, F., G{\'o}mez-Bombarelli, R., and Kutz, J.~N. (2019).
\newblock Data-driven modeling and learning in science and engineering.
\newblock {\em Comptes Rendus M{\'e}canique}, 347(11):845--855.

\bibitem[Montesinos~L{\'o}pez et~al., 2022]{montesinos_lopez_fundamentals_2022}
Montesinos~L{\'o}pez, O.~A., Montesinos~L{\'o}pez, A., and Crossa, J. (2022).
\newblock Fundamentals of artificial neural networks and deep learning.
\newblock In {\em Multivariate statistical machine learning methods for genomic prediction}. Springer.

\bibitem[M{\"u}ller and Guido, 2016]{müller_introduction_2016}
M{\"u}ller, A.~C. and Guido, S. (2016).
\newblock {\em Introduction to machine learning with Python: a guide for data scientists}.
\newblock O'Reilly Media, Inc.

\bibitem[Nassar et~al., 2005]{nassar_experimental_2005}
Nassar, S., El-Khiamy, H., Barber, G., Zou, Q., and Sun, T. (2005).
\newblock An experimental study of bearing and thread friction in fasteners.
\newblock {\em J. Trib.}, 127(2):263--272.

\bibitem[Nassar et~al., 2007]{nassar_effect_2006}
Nassar, S.~A., Ganeshmurthy, S., Ranganathan, R.~M., and Barber, G.~C. (2007).
\newblock Effect of tightening speed on the torque-tension and wear pattern in bolted connections.
\newblock {\em Journal of Pressure Vessel Technology}, pages 426--440.

\bibitem[Oberg, 2013]{oberg_machinerys_2004}
Oberg, E. (2013).
\newblock {\em Machinery's Handbook}.
\newblock New York: Industrial Press Inc.

\bibitem[Olejnik and Ayankoso, 2023]{olejnik_friction_2023}
Olejnik, P. and Ayankoso, S. (2023).
\newblock Friction modelling and the use of a physics-informed neural network for estimating frictional torque characteristics.
\newblock {\em Meccanica}, 58(10):1885--1908.

\bibitem[Ren and Sun, 2023]{ren_prediction_2023}
Ren, J. and Sun, X. (2023).
\newblock Prediction of ultimate bearing capacity of pile foundation based on two optimization algorithm models.
\newblock {\em Buildings}, 13(5):1242.

\bibitem[Roloff et~al., 2023]{spura_roloffmatek_2023}
Roloff, H., Spura, C., Fleischer, B., Wittel, H., and Jannasch, D. (2023).
\newblock {\em Maschinenelemente: Normung, Berechnung, Gestaltung}.
\newblock Springer-Verlag.

\bibitem[Ruder, 2016]{ruder2016overview}
Ruder, S. (2016).
\newblock An overview of gradient descent optimization algorithms.
\newblock {\em arXiv preprint arXiv:1609.04747}.

\bibitem[Sadouk et~al., 2020]{sadouk_robust_2020}
Sadouk, L., Gadi, T., and Essoufi, E.~H. (2020).
\newblock Robust loss function for deep learning regression with outliers.
\newblock In {\em Embedded Systems and Artificial Intelligence: Proceedings of ESAI 2019, Fez, Morocco}, pages 359--368. Springer.

\bibitem[Santry, 2024]{douglas_j_santry_demystifying_2024}
Santry, D.~J. (2024).
\newblock {\em Demystifying deep learning: An introduction to the mathematics of neural networks}.
\newblock Wiley.

\bibitem[Selvamuthu and Das, 2024]{selvamuthu_introduction_2024}
Selvamuthu, D. and Das, D. (2024).
\newblock {\em Introduction to Probability, Statistical Methods, Design of Experiments and Statistical Quality Control}.
\newblock Springer.

\bibitem[Steinhilper et~al., 2012]{steinhilper_konstruktionselemente_2012}
Steinhilper, W., Sauer, B., and Feldhusen, J. (2012).
\newblock {\em Konstruktionselemente des Maschinenbaus 1: Grundlagen der Berechnung und Gestaltung von Maschinenelementen}.
\newblock Springer-Verlag.

\bibitem[{VDI 2230}, 2014]{noauthor_vdi_2014}
{VDI 2230} (2014).
\newblock Blatt 2 -- systematische berechnung hochbeanspruchter schraubenverbindungen - zylindrische einschraubenverbindungen.

\bibitem[{VDI 2230}, 2015]{noauthor_vdi_2015}
{VDI 2230} (2015).
\newblock Blatt 1 -- systematische berechnung hochbeanspruchter schraubenverbindungen - zylindrische einschraubenverbindungen.

\bibitem[{VDI 2230}, 2024]{noauthor_vdi_2024}
{VDI 2230} (2024).
\newblock Blatt 3 -- systematische berechnung hochbeanspruchter schraubenverbindungen - zylindrische einschraubenverbindungen.

\bibitem[Wettstein and Matthiesen, 2020]{wettstein_investigation_2020}
Wettstein, A. and Matthiesen, S. (2020).
\newblock Investigation of the thread coefficient of friction when impact tightening bolted joints.
\newblock {\em Forschung im Ingenieurwesen-Engineering Research}, 84(1):55--63.

\bibitem[Wythoff, 1993]{wythoff_backpropagation_1993}
Wythoff, B.~J. (1993).
\newblock Backpropagation neural networks: a tutorial.
\newblock {\em Chemometrics and Intelligent Laboratory Systems}, 18(2):115--155.

\bibitem[Yadav et~al., 2015]{yadav_preliminaries_2015}
Yadav, N., Yadav, A., Kumar, M., Yadav, N., Yadav, A., and Kumar, M. (2015).
\newblock Preliminaries of neural networks.
\newblock {\em An introduction to neural network methods for differential equations}, pages 17--42.

\bibitem[Y{\i}ld{\i}r{\i}m et~al., 2019]{yildirim_development_2019}
Y{\i}ld{\i}r{\i}m, A., Akay, A.~A., G{\"u}la{\c{s}}{\i}k, H., {\c{C}}oker, D., G{\"u}rses, E., and Kayran, A. (2019).
\newblock Development of bolted flange design tool based on artificial neural network.
\newblock {\em Journal of Pressure Vessel Technology}, 141(5):051203.

\bibitem[Yu et~al., 2015]{yu_finite_2015}
Yu, Q., Zhou, H., and Wang, L. (2015).
\newblock Finite element analysis of relationship between tightening torque and initial load of bolted connections.
\newblock {\em Advances in Mechanical Engineering}, 7(5):1687814015588477.

\bibitem[Zeng et~al., 2020]{zeng2020analytical}
Zeng, J., Lu, W., and Paavola, J. (2020).
\newblock Analytical models for predicting the load-bearing strength of beam-to-column joints in a composite steel frame.
\newblock {\em Thin-Walled Structures}, 157:107100.

\bibitem[Zhong et~al., 2021]{zhong_prediction_2021}
Zhong, C.-j., Feng, R.-q., and Zhang, Z.-j. (2021).
\newblock Prediction of ultimate bearing capacity and structural optimization of aluminum alloy plate joints based on artificial neural network.
\newblock {\em International Journal of Steel Structures}, 21(5):1759--1774.

\bibitem[Zhou, 2021]{zhou_machine_2021}
Zhou, Z.-H. (2021).
\newblock {\em Machine learning}.
\newblock Springer nature.

\end{thebibliography}

\appendix

%\section*{Appendix}

%The \verb|\appendix| command signals that all following sections are
%appendices, and therefore the headings after \verb|\appendix| will be set
%as appendix headings.

%Note: All the figures, tables, equations, enunciations will be automatically
%numbered as A.1, A.2, etc. in the appendix part.
\end{Backmatter}

\end{document}